\documentclass[a4paper]{svproc}%
\usepackage{multirow}
\usepackage{booktabs}
\usepackage{stfloats} % For using figure* with tables
\usepackage{graphicx}
\usepackage{xcolor}
\usepackage[export]{adjustbox}
\usepackage{wrapfig}

\usepackage{amsmath}
\usepackage{amssymb}
\usepackage{mathtools}

\usepackage{tabularx}
\usepackage{algorithm}
\usepackage{color}
\usepackage{algorithmic}
\usepackage{amsmath}
\usepackage{hyperref}
\usepackage{url}
\usepackage{lipsum} % For dummy text

\renewcommand\footnotesize{\tiny} 

\begin{document}
\mainmatter              % start of a contribution
\title{Decentralized Vision-Based Autonomous Aerial Wildlife Monitoring}
%SkyNet for Wildlife: Decentralized Vision-Based Autonomous Aerial Monitoring
%Eyes in the Sky: Decentralized Autonomous Vision-Based monitoring for Wildlife 
\titlerunning{Decentralized Wildlife Monitoring}  % abbreviated title (for running head)
%                                     also used for the TOC unless
%                                     \toctitle is used
%
\author{Makram Chahine\inst{*,\dag,1,2} \and William Yang\inst{*,2} \and
Alaa Maalouf\inst{1,2} \and Justin Siriska\inst{1,3} \and Ninad Jadhav\inst{1,3} \and Daniel Vogt\inst{1,3} \and Stephanie Gil\inst{1,3} \and Robert Wood\inst{1,3} \and Daniela Rus\inst{1,2}}
\authorrunning{M. Chahine, W. Yang et al.} % abbreviated author list (for running head)
%
%%%% list of authors for the TOC (use if author list has to be modified)
% \tocauthor{Ivar Ekeland, Roger Temam, Jeffrey Dean, David Grove,
% Craig Chambers, Kim B. Bruce, and Elisa Bertino}
% %
\institute{$^{*}$ Equal contribution, \\
\inst{1}Project CETI, New York 10003, NY, USA, \\
\inst{2}CSAIL MIT, Cambridge MA 02139, USA, \\
\inst{3}SEAS Harvard University, Boston MA 02134, USA, \\
\inst{\dag}Corresponding author: \email{chahine@mit.edu}}

\maketitle              % typeset the title of the contribution

\begin{abstract}
Wildlife field operations demand efficient parallel deployment methods to identify and interact with specific individuals, enabling simultaneous collective behavioral analysis, and health and safety interventions. 
Previous robotics solutions approach the problem from the herd perspective, or are manually operated and limited in scale. 
We propose a decentralized vision-based multi-quadrotor system for wildlife monitoring that is scalable, low-bandwidth, and sensor-minimal (single onboard RGB camera). Our approach enables robust identification and tracking of large species in their natural habitat. We develop novel vision-based coordination and tracking algorithms designed for dynamic, unstructured environments without reliance on centralized communication or control. We validate our system through real-world experiments, demonstrating reliable deployment in diverse field conditions.
\begingroup
\renewcommand\thefootnote{}% suppress number
\makeatletter
\renewcommand\@makefntext[1]{\noindent\footnotesize #1}% remove indentation
\makeatother

\footnotetext{
This work was supported in part by the CETI Project, the Boeing Company, and the ONR Science of Autonomy program N00014-23-1-2354. The views and conclusions contained in this document are those of the authors and should not be interpreted
as representing the official policies, either expressed or implied, of the U.S. Government. The U.S. Government
is authorized to reproduce and distribute reprints for Government purposes
notwithstanding any copyright notation herein.}

\addtocounter{footnote}{-1}% keep counter unchanged
\endgroup
\end{abstract}

\begin{figure}
    \centering
    \vspace{-3.4em}
    \includegraphics[width=.6\linewidth]{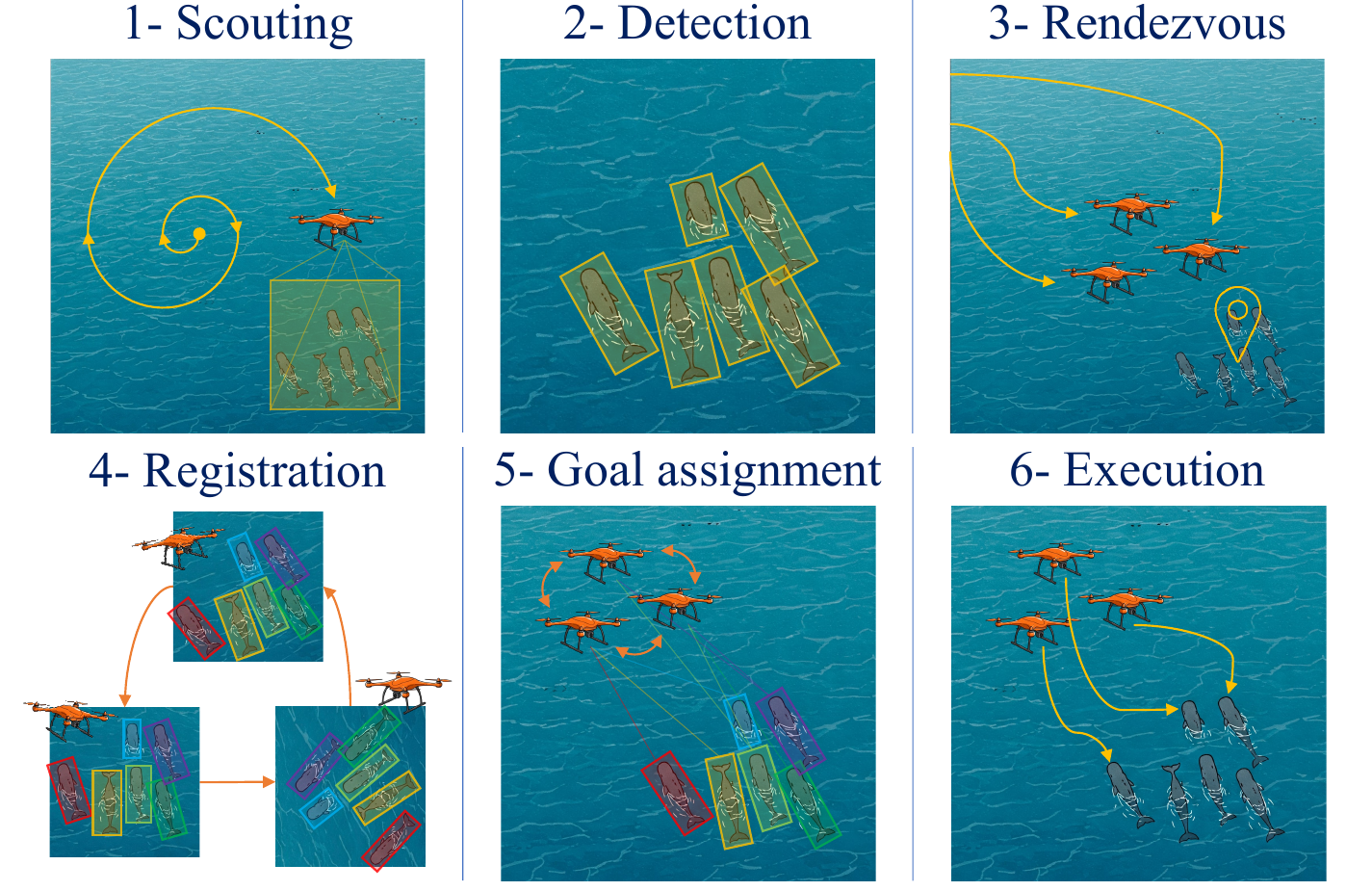}
    \vspace{-1em}
    \caption{Coordinated multi-quadrotor system for wildlife monitoring.}
    \label{fig:multiwhale_intro}
    \vspace{-1em}
\end{figure}
\clearpage

\section{Introduction}  

Monitoring wildlife at scale requires systems capable of parallel, individual-level observation across large environments. In particular, understanding species behavior and communication demands the ability to identify and track multiple animals simultaneously—critical for behavioral analysis, health monitoring, and intervention. Traditional monitoring strategies are often manual, centralized, or constrained in scale and adaptability.

We tackle the problem of detecting and tracking individual animals in the wild using autonomous aerial agents. Our solution targets scenarios such as marine mammal monitoring, where long-range detection and real-time coordination are vital. Using sperm whales as a running example, we introduce a fully autonomous, decentralized aerial system that relies solely on onboard RGB vision to detect species, register identities across agents, assign goals, and execute monitoring behaviors—entirely without centralized control or GPS.

This system introduces new modules for decentralized visual registration (Box-ICP), task allocation using graph neural networks, and end-to-end real-world deployment under communication and sensing constraints. Through real and simulated experiments, we validate its ability to perform robust group-level monitoring of large animals in unstructured, dynamic environments.

\section{Related Work}

\textbf{Vision-Based Detection and Navigation.} Object detection has seen dramatic progress over the past decade, beginning with two-stage frameworks like Faster R-CNN~\cite{ren2016fasterrcnnrealtimeobject}, which separate region proposal and classification, followed by efficient single-stage models such as SSD~\cite{Liu_2016} and YOLO~\cite{redmon2016lookonceunifiedrealtime}. More recent detectors incorporate transformer architectures, as in DETR~\cite{carion2020endtoendobjectdetectiontransformers}, or enhance inference speed and robustness through improved convolutional designs, such as in YOLOv8 and YOLOv12~\cite{reis2024realtimeflyingobjectdetection,tian2025yolov12}. These advances have enabled high-performance onboard detection suitable for real-time deployment on aerial robots.

In parallel, vision-based navigation has progressed from classical imitation learning systems~\cite{pomerleau1988alvinn,bojarski2016end} to modern end-to-end deep learning approaches~\cite{chib2023recent}, which achieve robust navigation in structured environments. However, these models often struggle under domain shift and lack interpretability. To address these issues, recent work has explored safety-aware learning~\cite{xiao2023barriernet} and robust generalization techniques, such as liquid neural networks~\cite{chahine2023robust}, Gaussian Splatting~\cite{quach2024gaussiansplattingrealworld}, and model-driven adaptation~\cite{wang2023learning,Yin2023,swift2023}. Moreover, emerging paradigms now integrate foundation models to enable flexible, instruction-conditioned policies~\cite{chahine2024flexendtoendtextinstructedvisual}, and scalable internet-scale perception backbones for real-time deployment~\cite{maalouf2024follow}.

\textbf{Goal Assignment.}  
The Linear Sum Assignment (LSA) problem~\cite{kuhnLSA} is a core formulation for multi-agent matching, with centralized solutions like the Hungarian algorithm widely used in robotics. Scalable extensions have been explored for team coordination~\cite{aziz2022taskallocationusingteam}, while decentralized alternatives address limited communication using Lyapunov-based methods~\cite{decgol}, distributed Hungarian approximations~\cite{dechun}, and iterative consensus protocols~\cite{dishun}. 

More recently, learning-based approaches have been developed to approximate LSA solutions. Deep neural networks have shown early promise~\cite{DNNLSA}, and graph neural networks (GNNs) offer a structured and scalable alternative for decentralized assignment. GLAN~\cite{liu2022glangraphbasedlinearassignment} frames the problem as bipartite edge selection, while subsequent works~\cite{tackling,gnndeclsa} demonstrate near-optimal decentralized goal allocation across varied team sizes and conditions, making them well suited to real-world swarm deployment.

\textbf{Visual Registration.}  
Registration techniques are fundamental for aligning observations across distributed agents. Classical approaches include the Iterative Closest Point (ICP) algorithm~\cite{chen1992object}, which aligns point clouds via nearest-neighbor matching and rigid transformations. Probabilistic extensions such as Gaussian Mixture Models (GMMs)~\cite{joshi1995problem} and Coherent Point Drift (CPD)~\cite{myronenko2010point} incorporate uncertainty and soft matching. Recent methods leverage learned representations~\cite{wang2019deep} or structural constraints such as graph or geometric features~\cite{agarwal2006bipartite}. These methods inform the design of our Box-ICP module, which incorporates structure-aware matching across bounding box edge sets in a decentralized visual registration setting.

\textbf{Consensus and Scalable Coordination.}  
Decentralized coordination frameworks have addressed robust consensus protocols~\cite{consensus}, decentralized formation flying \cite{ffgps1}, and scalable local game-theoretic planning~\cite{localchahine,intentionchahine}. These approaches enable agents to plan safe trajectories with minimal communication, informing our design of low-bandwidth, robust multi-agent control.

\section{Technical Approach}

The decentralized multi-agent wildlife monitoring pipeline proposed consists of six algorithmic components depicted in Figure \ref{fig:multiwhale_intro} and presented below:
\begin{enumerate}
    \item \textbf{Scouting:} Identify areas of potential \cite{ninadrdv} and search with spiral trajectories.
    \item \textbf{Detection:} Scan with vision model \cite{reis2024realtimeflyingobjectdetection} and share group's GPS coordinates.
    \item \textbf{Rendezvous:} Guide swarm to flock via decentralized formation control \cite{ffgps1}.
    \item \textbf{Registration:} Reach visual ID consensus on detection bounding boxes \cite{reis2024realtimeflyingobjectdetection}.
    \item \textbf{Goal Assignment:} Infer goals decentralized GNN-LSA \cite{gnndeclsa}.
    \item \textbf{Execution:} Perform visual monitoring via segmentation and tracking \cite{maalouf2024follow}.
\end{enumerate}
Our main contribution is the design of a novel, unified, cohesive vision-based system that encapsulates everything from scouting to individual task execution. A cornerstone of the pipeline is the visual detection module, trained on a curated dataset of real whale imagery obtained from field footage in the ocean as well as overhead shots of printed posters over land, augmenting the dataset for robust detection across different conditions. Detection is relied upon by the scout when looking for whales in a scene, but more importantly by the swarm of agents who use the produced bounding boxes as input to the core individual registration algorithm which enables fine grained goal assignment.\\
\noindent\textbf{Scouting, Detection, Rendezvous: } Scouting is achieved by deploying the scout agent to follow a predefined search pattern, e.g. growing spiral trajectory, around an area of high potential (either guessed by field teams or informed by sensing~\cite{ninadrdv}). The scout runs the detection model and, upon obtaining a robust whale presence signal (80\% of frames with detections in a buffer of recent images), is directed to visually track the filtered barycenter of all bounding boxes. Its current position is used as a target for the decentralized formation flying swarm.

Critical to our design are the \textit{registration} and \textit{goal assignment} modules, both of which operate under strict communication constraints using only visual input. To ensure scalability and reliability, we restrict inter-agent communication to a ring topology, where each drone communicates only with its two immediate neighbors.
In fact, we focus our technical contributions to vision-based decentralized registration and goal assignment. %We provide details on the technical approach adopted for these two critical decentralized components.

% \begin{wrapfigure}{r}{0.48\textwidth}
%     \vspace{-2.25em} % Adjust space to avoid collision with the text
%     \begin{minipage}{\linewidth}
%         \begin{algorithm}[H]
%         \caption{Decentralized Box-ICP}
%         \scriptsize % Reduce font size
%         \begin{algorithmic}[1]
%         \REQUIRE $n$ drones watching $m$ whales
%         \ENSURE Consistent registration among agents

%         \STATE \textbf{Initialize} transformations $\mathcal{T} \gets \{I\}_{1:n}$

%         \FOR{drone $d \in \{2,\dots,n\}$}
%             \STATE \textbf{Detect} bounding boxes $B_d$
%             \STATE \textbf{Receive} $B_{d-1}$ from predecessor
%             \STATE \textbf{Compute} $T_d$ using Box-ICP to align $B_d$ with $B_{d-1}$
%             \STATE \textbf{Update} $\mathcal{T}[d] \gets T_d \cdot \mathcal{T}[d-1]$
%         \ENDFOR

%         \texttt{\# Close the loop and verify consistency}
%         \STATE \textbf{Receive} $B_n$ at drone 1
%         \STATE \textbf{Compute} $T_1$ using Box-ICP to align $B_n$ with $B_1$
%         \STATE \textbf{If} $T_1 \cdot \mathcal{T}[n] \neq I$, \textbf{return} \texttt{False}
        
%         \STATE \textbf{Return} $\mathcal{T}$

%         \end{algorithmic}\label{alg:boxicp}
%         \end{algorithm}
%     \end{minipage}
%     \vspace{-2em}
% \end{wrapfigure}

\begin{algorithm}[t]
\caption{Box-ICP}
% \scriptsize % Reduce font size
\begin{algorithmic}[1]
    \REQUIRE Boxes $\mathcal{B}_1 = \{B_1^i\}_{i=1}^{N_1}$, $\mathcal{B}_2 = \{B_2^j\}_{j=1}^{N_2}$
    \ENSURE Transformation $T \in \mathbb{R}^{4 \times 4}$, Matching $\sigma: \{1,\ldots,N_1\} \to \{1,\ldots,N_2\}$
    
    \STATE Initialize $T \gets I_4$
    \WHILE{not converged}
        \STATE $C \gets 0 \in \mathbb{R}^{N_1 \times N_2}$
        \FOR{$i = 1$ to $N_1$}
            \FOR{$j = 1$ to $N_2$}
                \STATE Compute $D \in \mathbb{R}^{4 \times 4}$: edge-wise distances between $B_1^i$, $B_2^j$
                \STATE Solve LSA on $D$, set $C_{ij} \gets$ total match cost
            \ENDFOR
        \ENDFOR
        \STATE Solve LSA on $C$ for correspondence $\sigma$
        \STATE Form point sets $\mathcal{P}_1, \mathcal{P}_2$ from matched box edges
        \STATE Estimate $T'$ by least squares alignment on $(\mathcal{P}_1, \mathcal{P}_2)$
        \STATE $T \gets T' \cdot T$, update $\mathcal{B}_1 \gets T \cdot \mathcal{B}_1$
    \ENDWHILE
    \RETURN $T, \sigma$
\end{algorithmic}
\label{alg:boxicp}
\end{algorithm}

% \begin{wrapfigure}{r}{0.48\textwidth}
%     \vspace{-2.25em} % Adjust space to avoid collision with the text
%     \begin{minipage}{\linewidth}
%         \begin{algorithm}[H]
%         \caption{Box-ICP}
%         \scriptsize % Reduce font size
%         \begin{algorithmic}[1]
%             \REQUIRE Boxes $\mathcal{B}_1 = \{B_1^i\}_{i=1}^{N_1}$, $\mathcal{B}_2 = \{B_2^j\}_{j=1}^{N_2}$
%             \ENSURE \ \\ Transformation $T \in \mathbb{R}^{4 \times 4}$, \\  Matching $\sigma: \{1,\ldots,N_1\} \to \{1,\ldots,N_2\}$ \\ \
            
%             \STATE Initialize $T \gets I_4$
%             \WHILE{not converged}
%                 \STATE $C \gets 0 \in \mathbb{R}^{N_1 \times N_2}$
%                 \FOR{$i = 1$ to $N_1$}
%                     \FOR{$j = 1$ to $N_2$}
%                         \STATE Compute $D \in \mathbb{R}^{4 \times 4}$: edge-wise distances between $B_1^i$, $B_2^j$
%                         \STATE Solve LSA on $D$, set $C_{ij} \gets$ total match cost
%                     \ENDFOR
%                 \ENDFOR
%                 \STATE Solve LSA on $C$ for correspondence $\sigma$
%                 \STATE Form point sets $\mathcal{P}_1, \mathcal{P}_2$ from matched box edges
%                 \STATE Estimate $T'$ by least squares alignment on $(\mathcal{P}_1, \mathcal{P}_2)$
%                 \STATE $T \gets T' \cdot T$, update $\mathcal{B}_1 \gets T \cdot \mathcal{B}_1$
%             \ENDWHILE
%             \RETURN $T, \sigma$

%         \end{algorithmic}\label{alg:boxicp}
%         \end{algorithm}
%     \end{minipage}
%     \vspace{-2em}
% \end{wrapfigure}

\noindent\textbf{Individual Registration:} We design a scheme to align detected bounding boxes between successive agents, and achieve visual consensus on animal identities. Our Box-ICP algorithm is derived from the ICP approach, and adapted to incorporate constraints assigning boxes to individual agents, to align identities between pairs of agents (Algorithm \ref{alg:boxicp}). Indeed, after receiving box coordinates from its predecessor, an agent iteratively runs the following procedure until convergence: (i) Construct a matrix $D$ representing the \textit{distances between every pair of bounding boxes}, by determining the correspondence between the edges of every pair of boxes through solving LSA on the edge distance matrix. 
%by running LSA on the distance matrix between edges of the pair of boxes to find correspondence between the boxes edges. 
(ii) This is followed by solving another LSA on $D$ to minimize matching costs, and computing box correspondences (unlike classic ICP, which computes distances with all points without leveraging structural constraints). 
(iii) The optimal transformation is computed via least squares on all points in the point cloud. %to align the boxes. 

\noindent Along the ring communication graph, pairs of agents successively align their identity representations. Identity consensus is verified when returning to the first agent in the ring.
We assume all drones have downward-facing cameras, with all individuals contained in the images, which can be ensured by control of the swarm's formation and altitude.

\noindent\textbf{Goal Assignment:} We build on the Decentralized Graph Neural Network for Goal Assignment (DGNN-GA) framework \cite{gnndeclsa}, adapting for performance in low-communication-density settings and for situations where goals outnumber agents. Training data is generated by uniformly sampling agents and goals on a unit square, with agents connected cyclically based on their angular position with respect to the middle point. Goals available to each agent are capped in number and selected from their nearest neighbors. Euclidean distance is the metric of choice to define costs to goals, with the centralized Hungarian algorithm used to obtain an exact solution to label the training data. We generate 50k graphs for each $n_a$ and $n_g$ combination. We extend applicability to situations where goals outnumber agents by replacing the constraint requiring each goal to be assigned with a loss penalizing duplicate goal assignments, as discussed more formally below.

\noindent Let $A \in \mathbb{R}^{n_a \times n_g}$ be the soft assignment matrix (row $i$, denoted $A_i$, corresponding to the goal assignment probabilities for agent $i$). Let $\hat{A} \in \{0,1\}^{n_a \times n_g}$ be the hard assignment matrix obtained via row-wise $\arg\max$ over $A$.

\paragraph{1. Assignment Validity Loss.}  
Encourages each agent to be assigned to exactly one goal:
\[
\mathcal{L}_{\text{validity}} = \left\| \mathbf{1} - \sum_{i=1}^{n_a} \hat{A}_{i} \right\|_2 + \left\| \mathbf{1} - \left\| \hat{A}_{i} \right\|_2 \right\|_2
\]

\paragraph{2. Assignment Diversity Loss.}  
Encourages agents to select different goals by minimizing cosine similarity between their assignment vectors (with $\epsilon > 0$) :
\[
\mathcal{L}_{\text{div}} = \sum_{i=1}^{n_a} \sum_{j=i+1}^{n_a} 
\frac{A_{i} \cdot A_{j}}{
\max\left(\|A_{i}\|_2, \epsilon\right) \cdot \max\left(\|A_{j}\|_2, \epsilon\right)}
\] 

\paragraph{3. Supervised Optimization Loss.}  
Binary cross-entropy loss based on ground-truth labels $Y \in \{0,1\}^{n_a \times n_g}$:
\[
\begin{aligned}
\mathcal{L}_{\text{pos}} &= -0.9 \cdot \mathbb{E} \left[ Y \cdot \log(A + \epsilon) \right] \\
\mathcal{L}_{\text{neg}} &= -0.1 \cdot \mathbb{E} \left[ (1 - Y) \cdot \log(1 - A + \epsilon) \right] \\
\mathcal{L}_{\text{CE}} &= \mathcal{L}_{\text{pos}} + \mathcal{L}_{\text{neg}}
\end{aligned}
\]

\paragraph{Final Loss.}  
Total loss combining structure and supervision:
\[
\mathcal{L}_{\text{total}} = \alpha \cdot \left( \mathcal{L}_{\text{validity}} + \mathcal{L}_{\text{div}} \right) + (1 - \alpha) \cdot \mathcal{L}_{\text{CE}}, \quad \text{with } \alpha = 0.5
\]

\noindent\textbf{Execution: } Each agent uses the center point of the box corresponding to its assigned goal from the most recent frame available to it to initialize a segmentation mask as described in~\cite{maalouf2024follow}. Tracking of the obtained mask at a real-time running frequency of around 20 Hz enables robust monitoring of individual goals.

\section{Experiments}

We design a series of experiments, ranging from controlled offline evaluations to real-world multi-drone deployments, to validate each module of the system individually and in combination within the full pipeline.\\

%\subsection{Completed}
% \vspace{1em}
\noindent\textbf{Scouting and Detection:} Our visual detection model is trained on real drone footage of whales collected off the Dominica coast and flight tests over whale posters, enhancing dataset diversity. The training images are hand-labeled, with a total dataset of 1407 frames with a 90\% training and 10\% validation split for offline performance evaluation. Real-world testing involves a quadrotor flying a spiral trajectory over a soccer field with a whale poster. We validate the scout drone's ability to autonomously search, detect the poster, maintain position above it, and transmit its coordinates to a base station in real-time. We also run the module offline on real world data from the field.\\

\noindent\textbf{Detection and Registration:} To simulate various whale points of view, we synthesize 39 frames from aerial footage not in the training dataset and taken at three distinct timestamps, applying translations, rotations, and shearing, with an overlay of an ocean image to deal with empty pixels. The scene contains 9 whales.
We run detection, and the resulting bounding boxes serve as point clouds for Box-ICP matching. 
This test focuses on the algorithm's ability to handle noisy bounding box data and varying overlap ratios between frames.\\

\noindent\textbf{Goal Assignment:} GNN model evaluation uses large-scale runs with randomized agent and goal initializations to define graph conditions. 
Metrics measuring optimal cost proximity and redundant assignments assess models across varying agent/goal configurations (including more goals than agents). Since this is a high-level decision based solely on the cost matrix (pixel distance), models are evaluated at this abstraction level, independent of flight data. This evaluation specifically examines the GNN's performance in sparse communication graphs, mirroring the limitations of real-world drone swarms.\\

\noindent\textbf{Decentralized registration and assignment in the real world:} With a single drone capturing whale posters from various points of view, we test the validity of the decentralized vision-based detection, registration and goal assignment components in the real world and in real-time. \\

\noindent\textbf{Full Pipeline in the Real World:} The key test involves executing the entire pipeline onboard a fleet of drones in a real-world setting. For initial validation, we deploy two drones over a controlled environment containing six whale posters distributed across a soccer field. This setup engages all modules in the system—from autonomous scouting and detection to decentralized registration, GNN-based goal assignment, and multi-agent tracking—providing a complete, closed-loop evaluation of the approach.

\section{Results}
\begin{figure}[ht]
\centering
\vspace{-2.5em}
\includegraphics[width=.95\textwidth]{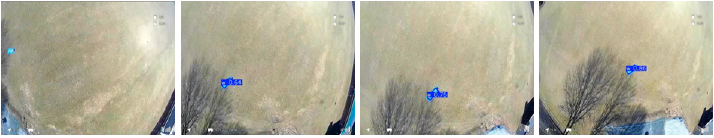}
\includegraphics[width=0.95\textwidth]{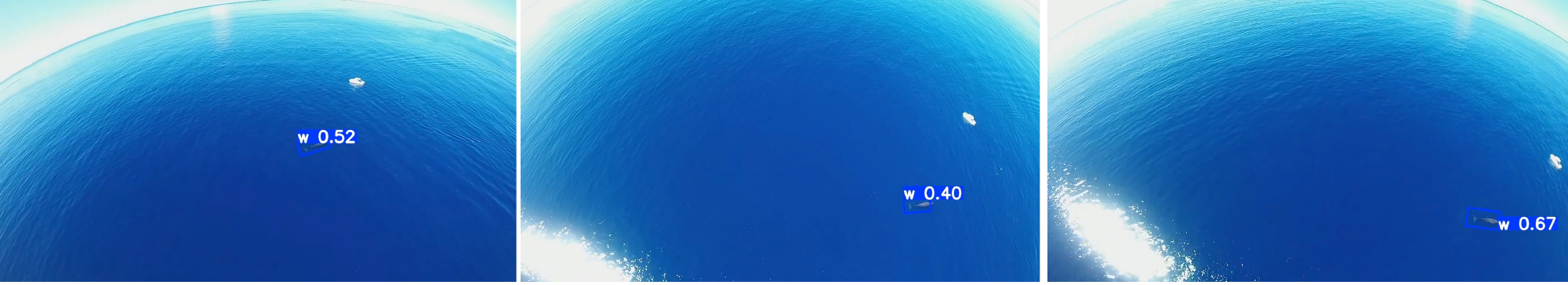}
\caption{Scouting detection runs. Top row depicts a full test with a whale poster where the columns from left to right corresponding to: whale appearance, first detection, drone centering on mean bounding box pixel position, transmitting coordinates.
Bottom row is an offline run of the detector on manual flight data from the field.}
\vspace{-2em}
\label{fig:detection}
\end{figure}

\noindent\textbf{Scouting and Detection:} The detector achieves satisfactory recall of 98.4\% and precision of 94.7\%, demonstrating its effectiveness in accurately identifying whales while minimizing false positives.
We tested autonomous search at realistic altitudes and exhibit sufficient robustness for real-world search missions. Indeed, relative to the poster size, test correspond to real-world flights at over 100m of altitude (modulo camera specs). Detection is not only accurate enough for locating whales but also serves as a visual reference for initial tracking. Figure~\ref{fig:detection} shows frames from detection runs.

Having trained multiple architectures including YOLOv8, YOLOv11, and DETR, we establish the former's supremacy in terms of performance on our dataset (see Table~\ref{tab:detect}).

\begin{table}[h]
\vspace{-1.5em}
\centering
\caption{Comparison of detection models on our whale dataset. OA: Overall Accuracy, FPR: False Positive Rate, FNR: False Negative Rate.}
\label{tab:detect}
\begin{tabular}{|l|c|c|c|}
\hline
\textbf{Model} & \textbf{OA (\%)} & \textbf{FPR (\%)} & \textbf{FNR (\%)} \\
\hline
YOLOv8s & \textbf{98.29} & \textbf{1.05} & \textbf{0.66} \\
YOLOv8n & 96.88 & 1.95 & 1.17 \\
YOLOv11n & 96.25 & 2.46 & 1.29 \\
YOLOv11s & 95.81 & 1.18 & 3.01 \\
DETR   & 58.08 & 9.70 & 32.22 \\
\hline
\end{tabular}
\vspace{-1em}
\end{table}

\begin{figure}[H]
\centering
% \vspace{-2em}
\includegraphics[width=.7\textwidth]{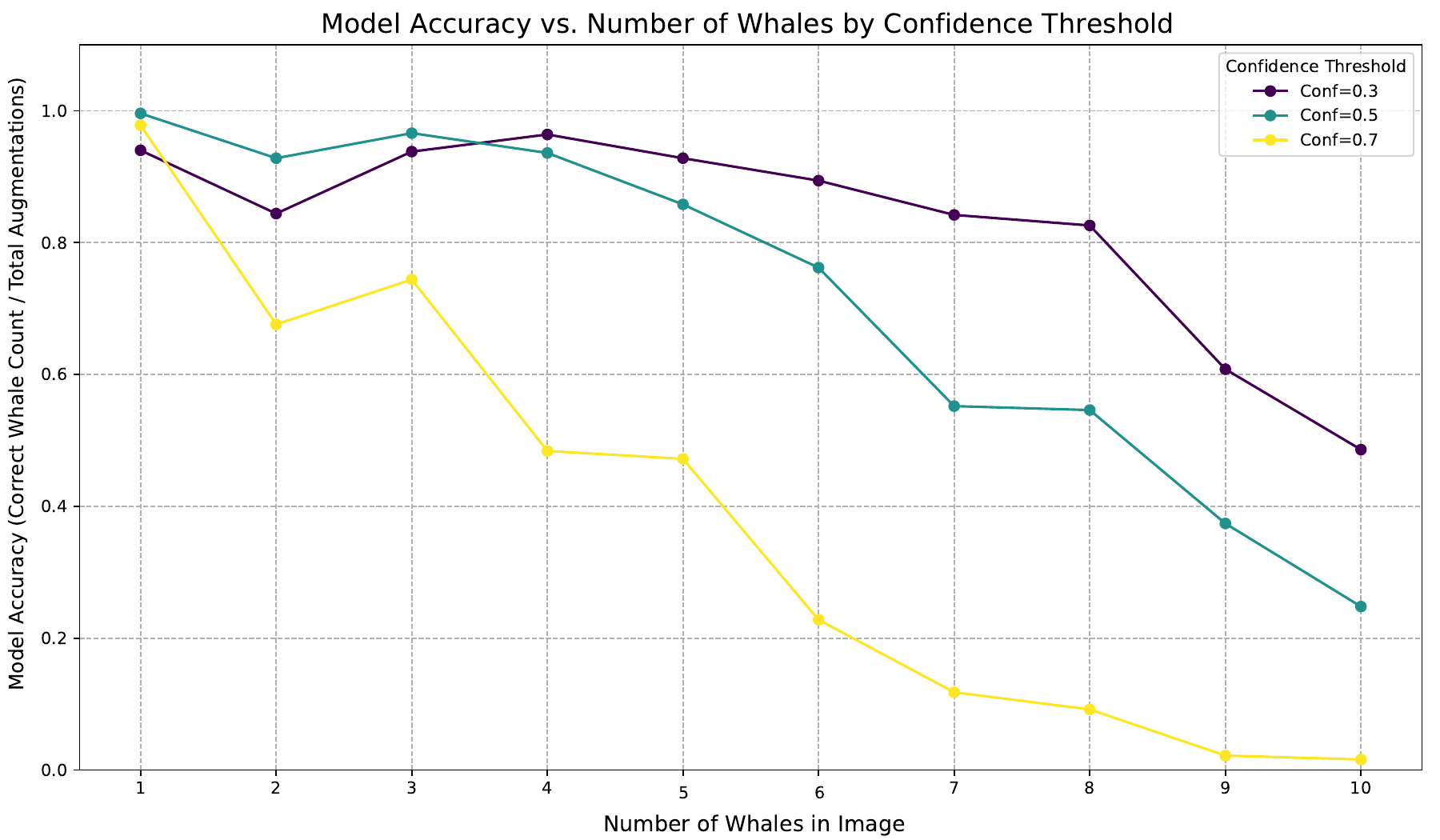}
% \vspace{-1em}
\caption{Detection success rate versus number of whales ($N=200$ per data point).}
\label{fig:detectsr}
% \vspace{-3em}
\end{figure}

\begin{figure}[ht]
\centering
\includegraphics[width=\textwidth]{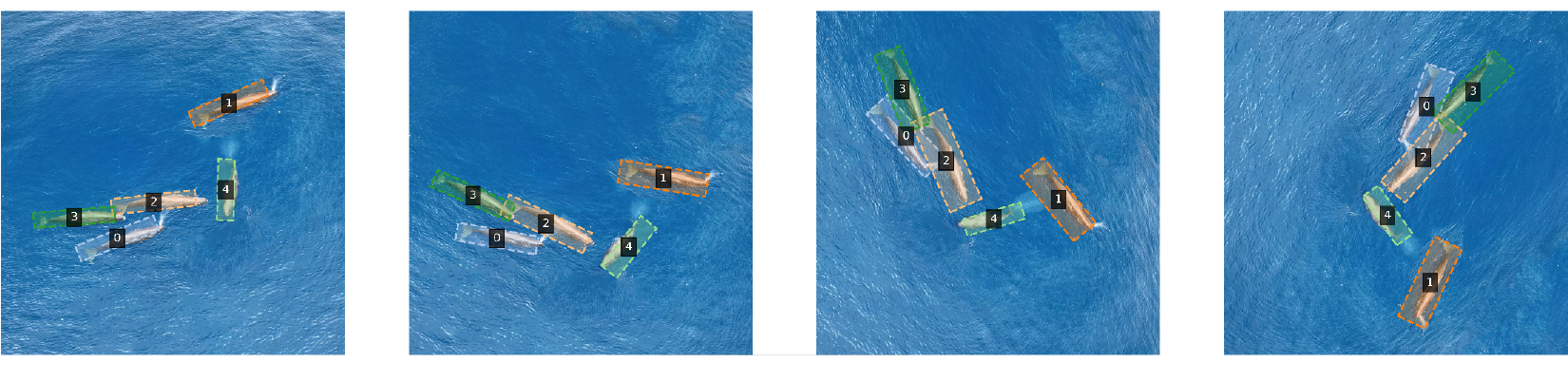}
\caption{Detection and individual registration with respect to the leftmost frame.}
\label{fig:register}
\end{figure}

On a custom evaluation dataset constructed from a single frame containing 10 whales, we generate various augmentations and examples with fewer goals using image editing tools to remove additional whales (examples for $n_g = 5$ in Figure~\ref{fig:register}). We evaluate the detector model's ability to correctly detect the exact number of whales in these images and provide the success rates for different numbers of whales and three detection confidence thresholds in Figure~\ref{fig:detectsr}.

\noindent\textbf{Detection and Registration:} 
On all 39 views of the 9-whale scene, our detector module detects the correct number of boxes, that appear properly positioned upon inspection.
We evaluate the pairwise success rate of Box-ICP in matching individuals for 200 pairs of frames from the 741 possible combinations. Every single assignment is correct, with examples depicted in Figure \ref{fig:register}.
In an $n$-drone ring network, the overall success rate follows  
$s(n) = s_{\text{p}}^n$,  
where the pairwise success rate is  
$s_{\text{p}} = s_{\text{det}}^2 \cdot s_{\text{reg}}$.  
The detector's reliability in multi-whale settings and the registration algorithm's perfect accuracy ensure scalability despite geometric success rate decay, making the system ready for real-world \textit{in situ} testing. 

\noindent\textbf{Goal Assignment:} 
% Instead of a surjective assignment loss, we introduce a differentiable agent assignment diversity loss to handle cases with more individuals than drones. For a graph $\mathcal{G}$, we define the diversity loss, $\mathcal{L}_{\text{div}} = \sum_{i=1}^{n_a} \sum_{j=i+1}^{n_a} \cos(\vec{a}_i, \vec{a}_j)$, 
% where $n_a$ is the number of agents, $\vec{a}_i \in \mathbb{R}^{n_g}$ is the assignment vector for agent $i$ with probabilities for $n_g$ goals, and $\cos(\vec{a}_i, \vec{a}_j)$ is the cosine similarity. Minimizing $\mathcal{L}_{\text{div}}$ encourages orthogonal assignment vectors, ensuring agents select distinct goals.
We evaluate models using an optimality score (\% of optimal assignments) and a diversity score (\% of non-overlapping solutions). Results averaged over 5000 random initializations with a GNN instance trained with 5 agents, 10 goals and 5 message passing rounds are summarized in Table \ref{tab:taskalloc}. We handle cases with $n_g < 10$ by adding $10-n_g$ high-cost ghost goals. We achieve both highly non-overlapping and close-to-optimal assignments while the number of goals remains close to the training setup (until $n_g = 7$). Having more goals than agents is a desirable setup for such a decentralized solution as even a single extra goal can help avoid overlaps around 80\% of the time. With a known fleet size, we can train a few suitable GNNs to offer robust performance over the spectrum of possible number of individuals encountered.

\begin{table}
\centering
\caption{Goal Assignment Results $n_a = 5$}
\label{tab:taskalloc}
\begin{tabular}{c|c|c|c|c|c|c}
\toprule
$n_g$ & $5$ & $6$ & $7$ & $8$ & $9$ & $10$ \\
\midrule
Optimality (\%) & 64.1 & 67.0 & 73.2 & 84.9 & 91.3 & 92.2 \\
Diversity (\%) & 34.3 & 78.8 & 91.2 & 93.3 & 93.6 & 94.6 \\
\bottomrule
\end{tabular}
\end{table}

% For the original non-differentiable version:

% $$\mathcal{L}_3 = \sum_{g=1}^{B} \mathbb{1}\left[\left|\text{unique}\left(\arg\max_j A_{g,i,j}\right)\right| \neq N_{\text{agents}}\right]$$

% \begin{wrapfigure}{r}{0.5\textwidth}
% \centering
% \vspace{-2.2em}
% \includegraphics[width=0.4\textwidth]{figs/all_drones_paths.pdf} % Replace with your image file
% \caption{In-simulation trajectories of a full pipeline run with 5 drones and 5 whales.}
% \label{fig:sim_traj}
% \vspace{-2em}
% \end{wrapfigure}
% \noindent\textbf{Full Pipeline Simulation:}
% Our Pybullet simulation setup allows us to test the integration of the different modules of the mission pipeline, with the trajectories of an entire simulation run depicted in Figure \ref{fig:sim_traj}. The scout drone scans a specified region, and upon detection of a pod of whales, flies towards the center of all detected bounding boxes. The scout shares the position with the tracking drones, which then fly in formation towards the pod, gathering around the scout drone. Cyclical Box-ICP on images obtained from each cameras' point of view.  
% Task assignment is handled by the decentralized GNN inference scheme, after which each drone attends to its goal. \\

\begin{figure}[t]
\centering
\vspace{-2em}
\includegraphics[width=\textwidth]{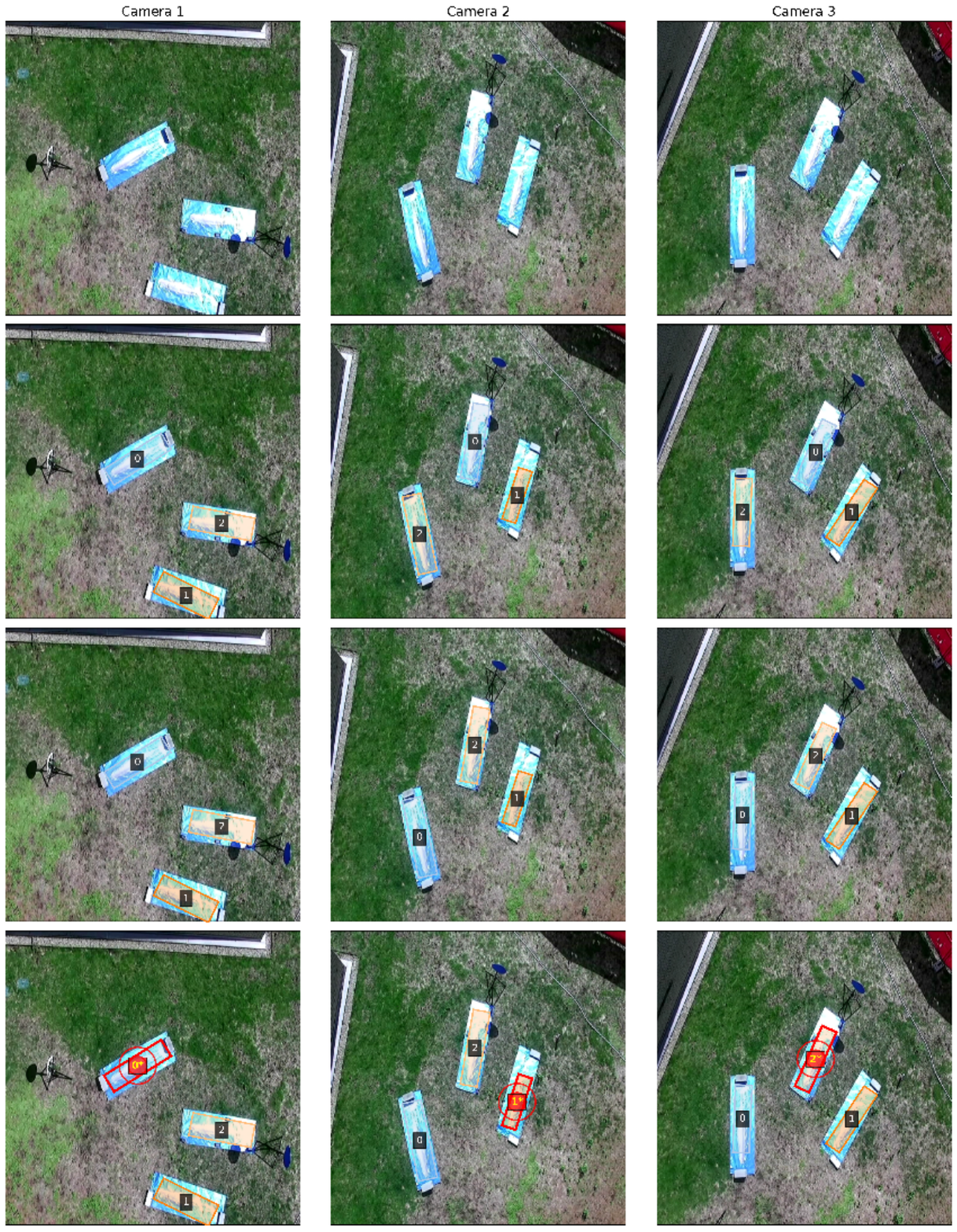}
\caption{Real-world real-time deployment of registration and goal assignment with 3 agents (per column) and 3 goals. The first row shows the raw frames, the second depicts the detections from each agent's point of view and with the initial identities, the third row presents the representations post Box-ICP alignment, and finally the bottom row shows the assigned goal to each agent in red.}
\label{fig:allinone}
\vspace{-3em}
\end{figure}

\noindent\textbf{Decentralized registration and assignment in the real world:} We demonstrate the system's viability using frames from 3 points of view to simulate the presence of 3 agents. The decentralized vision-based solution is run in real-time to detect whales, align representations and assign goals as depicted in Figure~\ref{fig:allinone}. \\

\noindent\textbf{Full Pipeline in the Real World:} We successfully deploy the entire pipeline onboard two drones flying over a soccer field with six whale posters placed as targets. A video demonstrating the complete step-by-step mission execution is included in the supplementary material and online via this \href{https://youtu.be/r3n0S-GpmWI}{link}. \\

\noindent\textbf{Communication Bandwidth:} Each whale is encoded as a four-coordinate box, consisting of $8$ values each $2$ bytes large. Each whale also has a corresponding index, which can be represented with $1$ byte since we do not expect more than $2^8$ whales in a given shot. Thus, for $N_w$ whales we expect to use $9N_w$ bytes. 

The drones also exchange hidden state vectors to facilitate the assignment of goals. Assuming our hidden state has dimension $d_h$, its size is then $4 \cdot d_h$ bytes. 

In our GNN, we use $d_h = 32$. Since we expect $N_w < 20$, our total message size is upper bounded by $32 \cdot 4 + 9 \cdot 20 = 308$ bytes. Assuming a radio connection that is at least 1MHz, our latency will be roughly $0.002$ seconds.

\section{Discussion}

 Our proposed pipeline, covering all mission aspects from autonomous scouting, robust detection, swarm gathering, individual registration, goal assignment, and monitoring execution, represents a novel approach to vision based real-time wildlife monitoring. The technical advancements in decentralized registration and GNN-driven task assignment, validated on real-world data, demonstrate a significant step forward in scalable and efficient multi-drone coordination for ecological applications. Furthermore, a full mission validation on hardware demonstrates the viability of the system as an efficient and effective real world solution for field scientists studying animal species in the wild.
 \clearpage

\bibliographystyle{spphys.bst} 
\bibliography{bib}

\end{document}